\documentclass[letterpaper]{article} 
\usepackage{aaai2026}  
\usepackage{times}  
\usepackage{helvet}  
\usepackage{courier}  
\usepackage[hyphens]{url}  
\usepackage{graphicx} 
\usepackage{natbib}  
\usepackage{caption} 
\usepackage{amsmath,amsfonts}
\usepackage{algorithmic}
\usepackage{algorithm}
\usepackage[justification=centering]{caption}
\usepackage[caption=false,font=normalsize,labelfont=sf,textfont=sf]{subfig}
\usepackage{textcomp}
\usepackage{url}
\usepackage{verbatim}
\usepackage{graphicx}
\usepackage{multirow}
\usepackage{makecell}
\usepackage[english]{babel}
\usepackage{booktabs}
\usepackage{pifont}
\usepackage{xcolor}
\usepackage{fontawesome}

\usepackage{newfloat}
\usepackage{listings}
\DeclareCaptionStyle{ruled}{labelfont=normalfont,labelsep=colon,strut=off} 
\floatstyle{ruled}
\newfloat{listing}{tb}{lst}{}
\floatname{listing}{Listing}
\title{Tempo-R0: A Video-MLLM for Temporal Video Grounding through Efficient Temporal Sensing Reinforcement Learning}
\author{
    Feng Yue,
    Zhaoxing Zhang,
    Junming Jiao,
    Zhengyu Liang,
    Shiwen Cao,
    Feifei Zhang,
    Rong Shen
}
\affiliations{
    Li Auto Inc.\\

    Beijing, 101303 China\\
}

\begin{document}

\maketitle

\begin{abstract}
Temporal Video Grounding (TVG), which requires pinpointing relevant temporal segments from video based on language query, has always been a highly challenging task in the field of video understanding. Videos often have a larger volume of information and redundancy than texts or images. Models should present comprehensive understanding of the whole video to accurately retrieve query-relevant clips. We thus propose Tempo-R0: a Video Multimodal Large Language Model (Video-MLLM) for the temporal video grounding task via multimodal temporal sensing reinforcement. Specifically, during the preprocessing stage of our pipeline, we employ Self-adaptive Attention Allocation (\textbf{SAA}) method based on frame content variation to efficiently use the MLLM's limited attention. The Explicit Timestamp-modal Aligned (\textbf{ETA}) method is also utilized to strengthen our model's capability to perceive the boundaries of events in the video. In the fine-tuning part of our pipeline, we creatively apply Partial Irrelevance Refusing-based Group Relative Policy Optimization (\textbf{PIR-GRPO}) in TVG area to foster model's temporal reasoning from not only accepting relevant video-query pairs but also refusing irrelevant ones. Experiments demonstrate that our method accomplishes a notable advantage over SOTA solutions by around 3.5\% on both the original QVHighlights testbench and its corrected version with more reasonable ground truth annotations.
\end{abstract}

\section{Introduction}
Multimodal large language models (MLLMs) have gradually emerged as the core of solutions to problems such as image and video understanding \cite{chatgpto1, chatgpto3o4mini, claude4sonnet, liu2023improved, li2022blip, li2023blip2, xue2024xgenmm, chen2025blip3ofamilyfullyopen, wang2024qwen2vlenhancingvisionlanguagemodels, bai2025qwen25vltechnicalreport}. As the parameter size of MLLMs scales up and the design of these MLLMs are rationalized, the corresponding gain on understanding tasks including summarization, captioning and question answering has been continuously realized. However, for Temporal Video Grounding (TVG) task which requires the capabilities not only to accurately perceive temporal boundaries between different semantic segments within videos but also to sense the cross-modal relevance between these video segments and retrieval queries. As a result, these pre-trained MLLMs nevertheless fail to achieve the same level compared to the aforementioned understanding tasks.

This shortcoming can be attributed to the following points according to our analysis:

\begin{itemize}
    \item The conflict between the redundancy of video information and the context-length constrain of MLLMs makes both these semantically independent events and their temporal boundaries difficult to be precisely perceived.
    \item The focus mismatch between video understanding tasks such as captioning or summarization applied in the pre-training stages in current MLLMs and the TVG tasks that they lack.
    \item Trainable datasets for TVG tasks are generally more difficult to obtain and corresponding data augmentation manipulations are relatively complex.
\end{itemize}

In light of this, we propose Tempo-R0, a Video-MLLM designed for TVG task via multimodal temporal sensing reinforcement. Established upon the pre-trained Qwen2-VL-7B model as the backbone MLLM, the multimodal temporal sensing reinforcement in Tempo-R0 is mainly realized in two aspects: methods to enhance temporal perception capabilities through efficient application of limited MLLM context and explicit timestamp injection for the TVG task, and Reinforcement Learning with Verifiable Reward (RLVR) based two-staged Reinforced Fine-Tuning (RFT) training policy conducted on a delicately curated dataset containing mixture of not only in-distribution, normal but also out-of-distribution, pure irrelevant training data. These improvements mentioned are described in Figure \ref{tr0-i0}:

\begin{figure}[!ht]
\centering
\captionsetup{justification=raggedright, singlelinecheck=false}
\includegraphics[width=0.97\columnwidth]{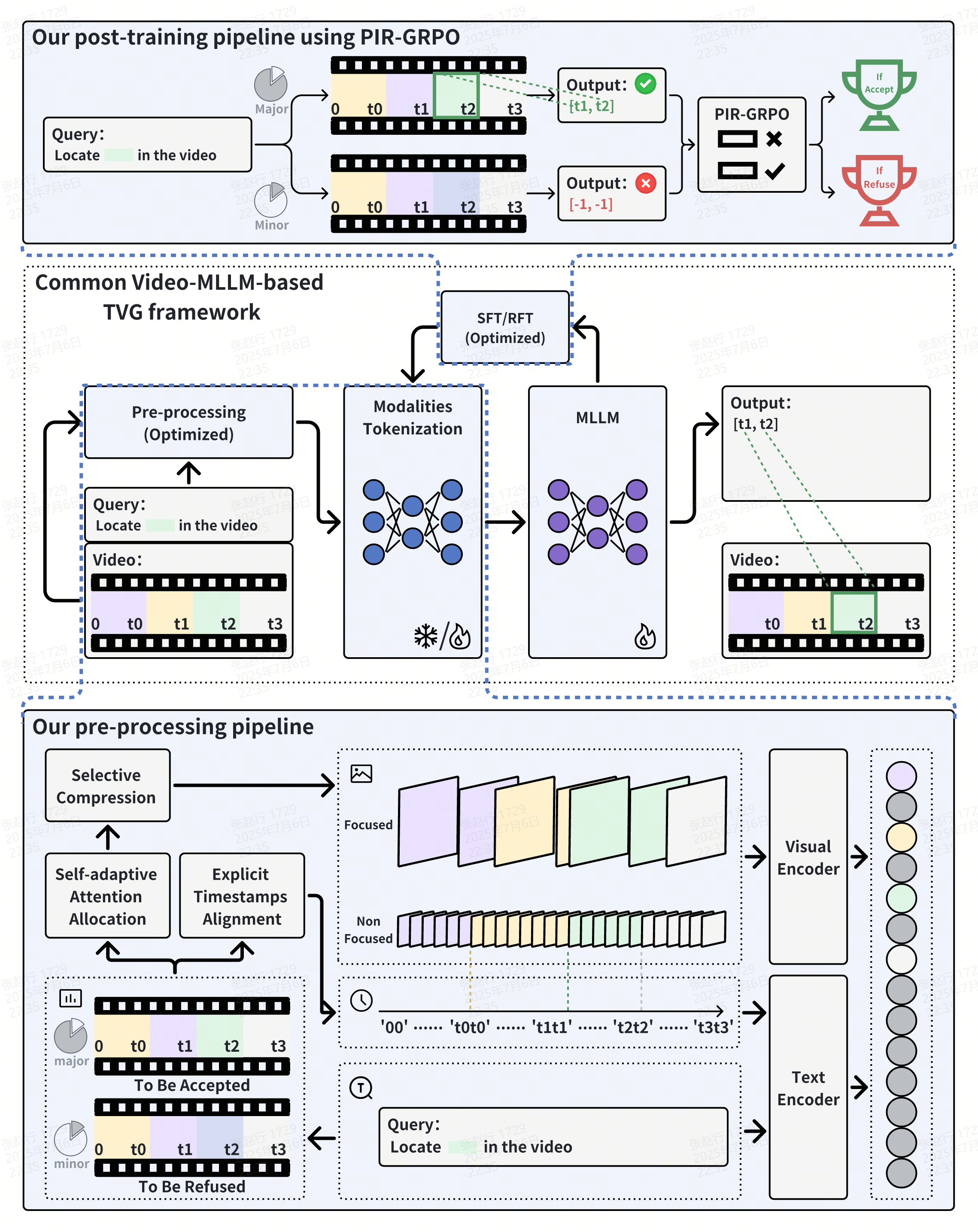} 
\caption{The major improvements realized by our Tempo-R0 compared to other Video-MLLMs.}
\label{tr0-i0}
\end{figure}

In summary, our major contributions are listed as follows:
\begin{itemize}
    \item \textbf{Self-adaptive Attention Adjustment (SAA):} To efficiently use the limited contextual token of MLLM, we introduce visual criterion to detect frames with large-scaled content fluctuation among temporally consecutive sampled frames, thus retaining these information-rich frames for greater possibility of moment segmentation and accurate boundaries perceiving.

    \item \textbf{Explicit Timestamp Alignment (ETA):} By treating timestamps as an independent modality, we explicitly feed them into the model along with other modalities after format alignment in order to enhance the temporal localization capability of MLLMs.

    \item \textbf{Partial Irrelevance Refusing-based Group Relative Policy Optimization (\textbf{PIR-GRPO}):} By appropriately adding pure irrelevant videos training data containing no relevant information to queries and corresponding adaption on the original GRPO method, we teach the MLLM to learn effective refusion and furthermore improve MLLM's average performance for TVG tasks in our two-staged RFT fine-tuning process.

    \item \textbf{A More Reasonable Testbench:} Through manual verification, we rectify the ground truth annotations of the test split of the QvHighlights \cite{lei2021qvhighlights} dataset and provide the corrected QvHighlights dataset, which is more consistent with human perception. In subsequent work, we will use it together with the original QvHighlights to evaluate the TVG capabilities of various approaches.
\end{itemize}

\section{Related Works}
Currently, there are primarily two categories of frameworks for the TVG task. The first is based on small-to-medium-scale networks built on transformer networks to perform multimodal feature compression and fusion, and generate the starting and ending timestamps of the video clips that are most relevant to the query using designed output heads \cite{Gao_2017, moon2023correlationguided, moon2023query, aleks2024saliencyguided, Jang_2023}. The second category of framework often relies on MLLMs to accomplish this cross-modal match and directly output the same results through prompting, pre-processing and fine-tuning.

\textbf{Non-Video-MLLM:} This type of method usually applies complex network architectures featuring Convolutional Neural Network (CNN) based or transformer-based network modules to perform multimodel inputs feature encoding, alignment, compression, fusion and results decoding. Early solutions under this framework \cite{Gao_2017} generally implement the aforementioned modules using CNN-like networks, while later ones \cite{moon2023correlationguided, moon2023query, aleks2024saliencyguided, Jang_2023} adopt transformer-based network structure to improve feature fusion through a wider receptive field from more training data. In terms of training, most Non-Video-MLLM methods employ an end-to-end training style as the whole network is directly supervised by regression loss. Tricks such as online irrelevant sample mining are also introduced in some approaches, enabling contrastive loss supervision to enhance the model's performance.

\textbf{Video-MLLM:} Compared to Non-Video-MLLM methods, Video-MLLM-like frameworks \cite{meinardus2024chrono, yu2023selfchained, li2025videochatr1, wang2025timer1, wang2024hawkeye} replace complex networks with pre-trained MLLMs, followed by fine-tuning phase on TVG task-related datasets in order to transfer the acquired knowledge and reasoning capability from the multimodal understanding area to the TVG area. Video-MLLM offers the following advantages over Non-Video-MLLM methods:

\begin{itemize}
    \item Temporal information can be soundly integrated into the pre-training process of the MLLM such as Qwen2.5's 3D temporal-spatial union position encoding, thus assigning these MLLMs inherent temporal perception capability from the scratch and making the TVG-oriented fine-tuning process followed easier and smoother.
    \item According to research \cite{aleks2024saliencyguided}, joint training with other cross-modal tasks such as video summarization or question answering is able to improve performance on TVG task. And Video-MLLM framework supports this as they are designed and trained to be fed with interleaved multimodal data in both TVG and other cross-modal tasks.
    \item Efficient RFT-based training methods can be conveniently merged into the Video-MLLM framework to strengthen its reasoning and temporal perception capabilities to handle TVG task.
\end{itemize}

In light of this, we propose the Tempo-R0 solution based on the Video-MLLM framework. Specifically, to address the contradiction between the limited context capacity and the rich information videos contain, we enhance the model's ability to understand the relevance of multimodal content and perceive the temporal boundaries of multimodal events through the \textbf{SAA} and \textbf{ETA} modules in the preprocessing stage, together with the data-driven \textbf{PIR-GRPO} in the RFT phase.

Our \textbf{SAA} is based on the Optimized Transportation (OT) based frame variations detector \cite{peyré2025optimaltransportmachinelearners} as it selects frames with obvious or relatively subtle content changes for high definition sampling and those with smaller change magnitudes for low resolution sampling. This enhances the possibility that visual tokens regarded as boundary frames of each independent semantic segment in the video gain more attention of the MLLM. Unlike \cite{Feichtenhofer_2019}, we mix these sampled images of different resolutions instead of the separation first fusing later policy, with the expectation of retaining the uniformity of extracted features.

In addition to the involvement of temporal information through position encoding \cite{wang2024qwen2vlenhancingvisionlanguagemodels, bai2025qwen25vltechnicalreport}, some solutions like \cite{meinardus2024chrono} provide a similar explicit timestamp injection method to our \textbf{ETA} to strengthen the temporal sensing performance in TVG. However, it tokenizes timestamps in the form of single discrete token. This deprivation of their original continuity will possibly increase the difficulty for MLLM to precisely align sampled temporal frame sequences with content comprehension in terms of the event boundary detection. We prove this via quantized comparison through ablation.

For our development of \textbf{PIR-GRPO} in Tempo-R0, contemporaneous research by \cite{Flanagan_2025} also introduces pure irrelevant samples with irrelevant video and query pairs. Compared to their Supervised Fine-Tuning (SFT) only training policy and the modification on the the output header, we focus on the reasoning development that viewing effectively refusing as a complement to the improvement of correct acceptance ability. We also prove in the ablation study section that this gain can be achieved at no cost of network modification but by means of proper proportion of pure irrelevant samples in our two-staged RFT process.

Furthermore, during our testing process, we observe and correct some ground-truth annotations inconsistent with common sense in the QvHighlights evaluation data set. We will present the performance of Tempo-R0 on the QvHighlights dataset before and after correction in the following experiments respectively.

\section{Methodology}
\begin{figure*}[!ht]
\centering
\captionsetup{justification=raggedright, singlelinecheck=false}
\includegraphics[width=2.0\columnwidth]{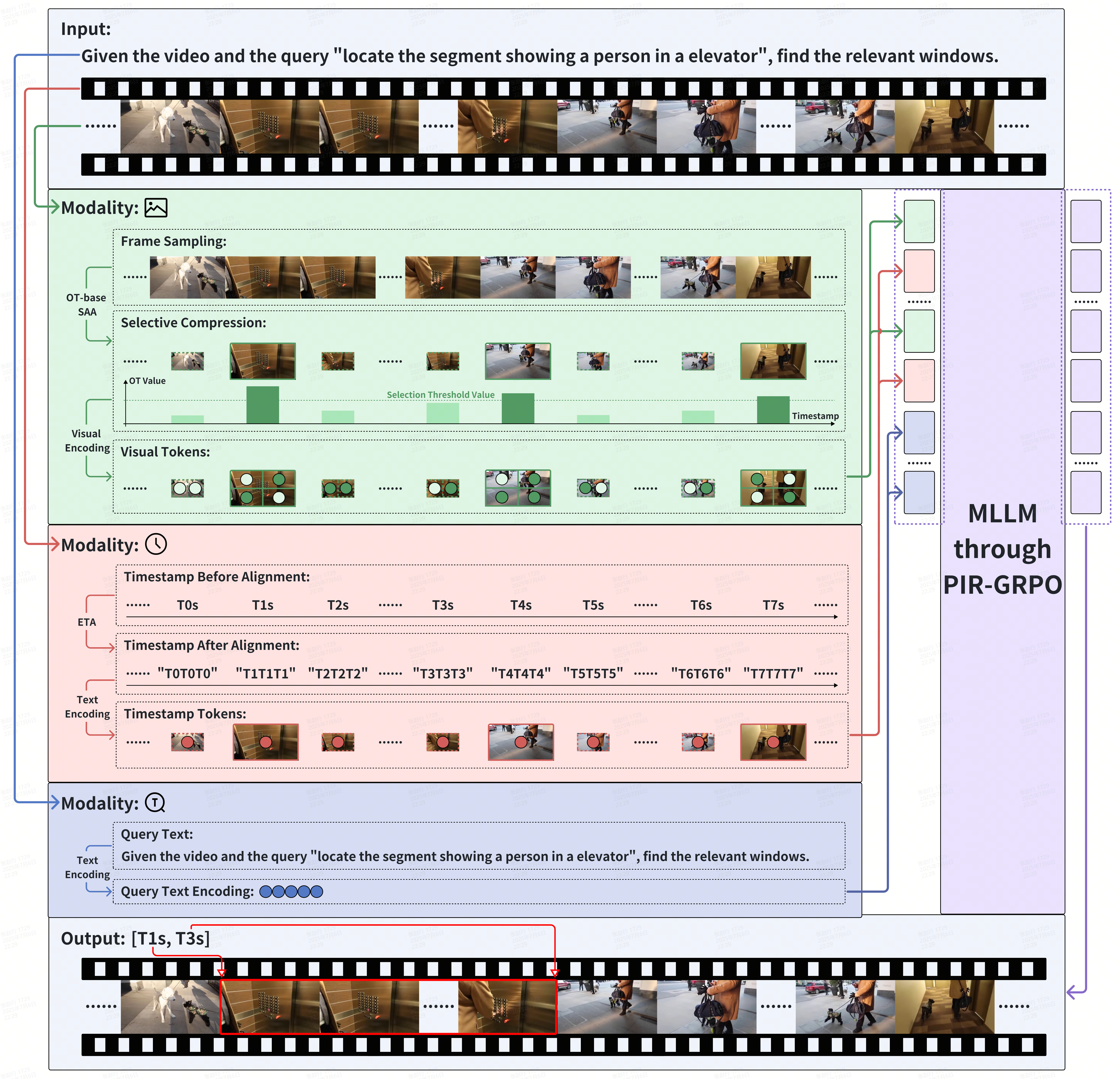}
\caption{In the inference pipeline of Tempo-R0, the inputs are decomposed into three independent modalities for separate processing during the pre-processing stage. For the visual modality, the sampled frames are selectively compressed and tokenized through OT-based evaluator in \textbf{SAA}. For the timestamp modality, Tempo-R0 converts each frame’s timestamp from a floating point into text and ensures digit character number equalization via zero-padding in \textbf{ETA}. The processed timestamp tokens and visual tokens are interleaved offered to the MLLM with properly-designed prompts. The MLLM directly outputs the starting and terminating timestamps of the potential matched clips.}
\label{tr0-i1}
\end{figure*}

The ability of Tempo-R0 to detect boundaries of relevant video clips during the inference phase is mainly contributed to the aforementioned \textbf{SAA} and \textbf{ETA} modules. We visually demonstrate how they are integrated with the MLLM backbone and work collaboratively in our pipeline in Figure \ref{tr0-i1}. During our RFT process, the pre-trained MLLM is first trained through SFT on all the parameters with datasets containing normal In-Domain (ID) and Out-Of-Domain (OOD) data. After the SFT stage, the MLLM is then trained by the proposed \textbf{PIR-GRPO} using curated datasets that include irrelevant and normal training data in the domain. We will discuss these modules and in detail in the following sections.

\subsection{\textbf{Self-adaptive Attention Allocation (SAA) for Visual Modality}}
Localization ability of event boundaries in video plays a significant role in TVG task, which is affected by the limited context memory length of MLLM according to previous elaboration. Take a $T$-second video as an example, sampling at $f$ fps will generate $F=\lfloor T/f \rfloor$ frames. The MLLM holding $L$-token visual context budget is able to only allocate $\lfloor F/L \rfloor$ tokens for each input frame image. This corresponds to approximately a rather low resolution of 224×140 in the case of Qwen2-VL-7B's configuration, which cannot fully represent visual details and weaken MLLM's temporal localization capability. Meanwhile, information redundancy sourced from sampled frames sequence might lead to disturbance of precise events boundary detection. As a result, the degree of salience of these potential boundary frames with drastic content changes are weakened by other redundant ones. Therefore, we primarily utilize \textbf{SAA} to allocate more tokens to images with drastic information changes to increase their significance in the eyes of MLLM. 

\begin{figure}[!htbp]
\centering
\captionsetup{justification=raggedright, singlelinecheck=false}
\includegraphics[width=0.97\columnwidth]{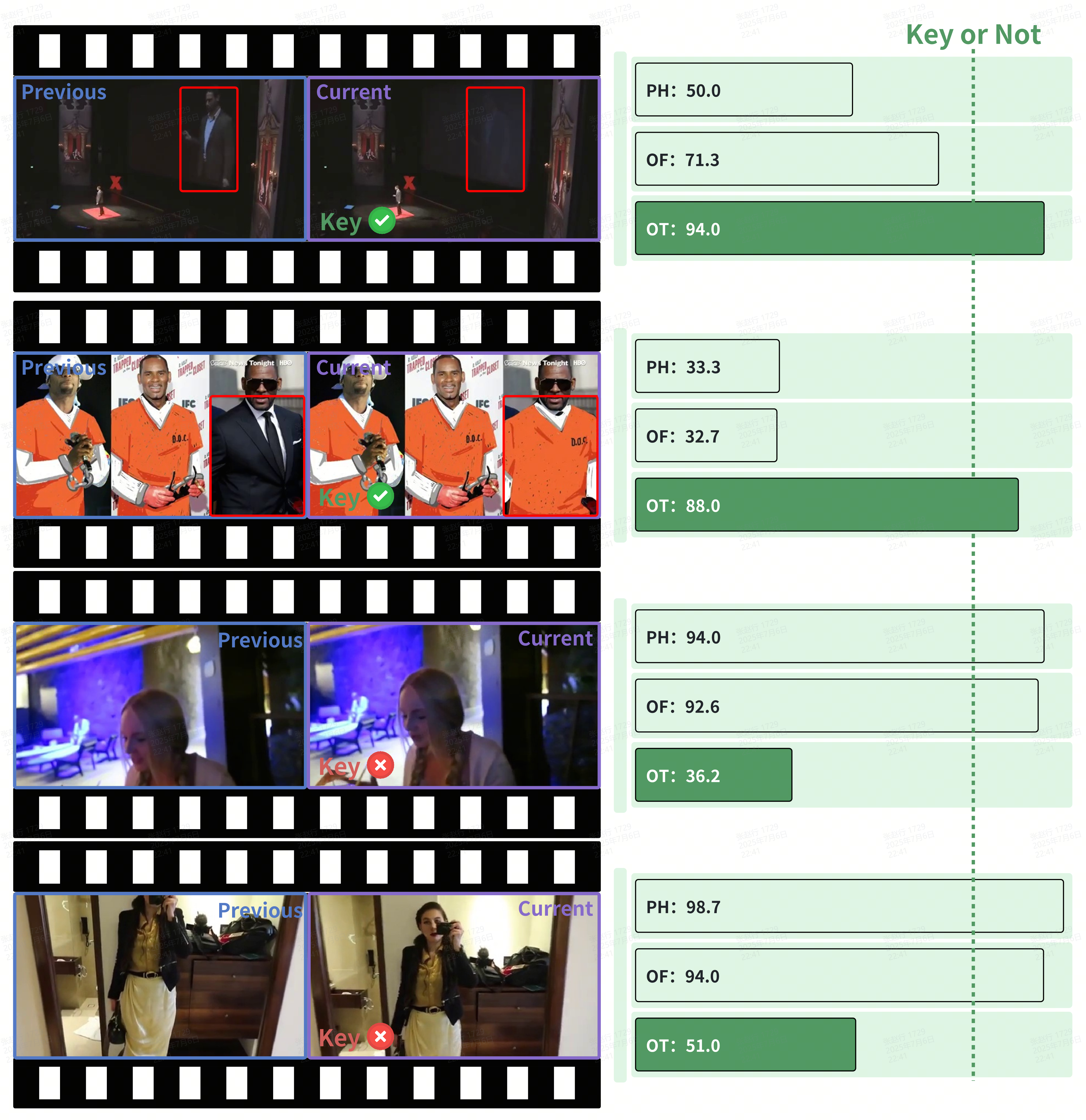}
\caption{The advantage of OT-based methods in key image perception from image sequences in \textbf{SAA}. Content changes are marked with red boxes in the identified key images. The scores from different methods are normalized.}
\label{tr0-i2}
\end{figure}

We only utilize the hue-channel component of images that are converted to the HSL color space, trying to neglect the visual content fluctuation impacted by different illumination. We then adopt the OT-based method to quantify the degree of content variation between adjacent frames in the sampled sequence. Compared to PHash (PH) and Optical Flow (OF), OT is generally more sensitive to the occurrence of new objects or exteriors in image sequences, as Figure \ref{tr0-i2} demonstrates. We assign higher weights to these key images through scaling factors to obtain more visual tokens after they are encoded. The complete \textbf{SAA} processing of one frame is represented in \eqref{eq1}:
\begin{equation}
Token_{v_i}=Visual\_Encode(Resize(F_{i},R)) \text{ } F_{i} \in F
\label{eq1}
\end{equation}
The $Resize()$ in \eqref{eq1} represents the resizing of the image using the downsampling scale $R$ that satisfies the following: 
\begin{equation}
\label{eq2}
R=\begin{cases} R_l, & \text{\it{if}\quad} OT(H(F_{i}))<T_{Key} \\ R_s, & \text{\it{otherwise} } \end{cases} \textit{and } R_l>R_s
\end{equation}
The extracted Hue-channel frame data $H(F_i)$ are evaluated by $OT()$. Frames with evaluation scores above the preset threshold value are viewed as key frame and are downsampled with smaller scaling factor than others.

\subsection{\textbf{Explicit Timestamp Alignment (ETA) for Timestamp Modality}}
MLLMs usually embed temporal information in an implicit style, such as 3D positional encoding, that makes it difficult for them to precisely connect events to timestamps, as shown in Figure N.
According to \cite{meinardus2024chrono}, the TVG task is transformed from the task of generating timestamps to the task of retrieving timestamps, reducing the difficulty of the model to perform temporal mapping of content. Unlike \cite{meinardus2024chrono, wang2024qwen2vlenhancingvisionlanguagemodels}, we align the timestamps of the sampled frames $S_{ts_i}$ in text format $\hat{S_{ts_i}}=Align(S_{ts_i})$ such that the original timestamps containing different digits have the same token length after alignment. The experiments show that this aligned approach releases the burden on the MLLMs to understand the timestamp modality, leading to an increase in final accuracy. The \textbf{ETA} process can be described as:
\begin{equation}
\label{eq3}
Token_{ts_i}=Text\_Encode(Align(I_{ts_i}))
\end{equation}

We then interleave the aligned timestamp tokens obtained from \eqref{eq2} with the visual tokens from \eqref{eq1} and \eqref{eq2} in the \textbf{SAA} and feed them to the MLLM with the TVG task instruction prompt to generate the output. Figure \ref{tr0-i1} illustrates how \textbf{SAA} and \textbf{ETA} cooperate in the inference pipeline of Tempo-R0. 

\subsection{Partial Irrelevance Refusing GRPO (\textbf{PIR-GRPO}) in Two-Staged Reinforced Fine-Tuning}
To further enhance the temporal reasoning ability of Tempo-R0, we adopt a two-stage fine-tuning approach including the SFT stage and the RLVR-based RFT stage on the MLLM. 

In the SFT stage, we first fine-tune the pre-trained MLLM using the ID training part of these datasets in order to assigning it relevant moment retrieval capability. For the purpose of generalization, we augment some OOD data. But we only select these with highly similar content to prevent confusing the MLLM with excessive diversity.

We exclusively leverage the ID training dataset and employ the GRPO in the RFT stage. Built upon the SFTed MLLM, it allows the MLLM to further enhance its reasoning ability through the response tendency shift from low-scored to high-scored answers to match relevant video clips to queries. Following the same logic, the response tendency shift from high-scored to low-scored answers to refuse irrelevant video-query matches should also be focused. In short of this reasoning, the model will always predict video intervals as the response to irrelevant query. We analyze that in this condition the MLLM tend to seek reward by making arbitrary guesses and this degradation will also impact the reasoning in relevant cases. Therefore, we propose the \textbf{PIR-GRPO} through the modification of the precision reward function with curated training data to develop this irrelevance rejection capability of MLLM.

Specifically, the reward functions of our \textbf{PIR-GRPO} are composed of the format reward function $R_{f}$ and precision reward function $R_{p}$. Each of Tempo-R0's matched video clip $Prd_{i}, i \in 0,1,...$ are expressed as $[St_{i},Ed_{i}], i \in 0,1,...$ where $St_{i},Ed_{i}$ represents the starting and ending timestamps. Therefore, $R_{f}$ are defined in \eqref{eq4}:
\begin{equation}
\label{eq4}
R_{f}=\begin{cases} 1, & \text{\it{if}\quad contains\space} ``[St_{i}, Ed_{i}]" \\ 0, & \text{\it{otherwise} } \end{cases}
\end{equation}
The $R_{f}$ are decomposed into two items $R_{tvg}$ and $R_{pir}$ as \eqref{eq5} shows, in which the expression $\mathbb{I}_{Prd_{i}=\emptyset}$ means that the MLLM will be rewarded 1.0 for correct rejection. $R_{tvg}$ are common precision reward function in TVG task for the relevant video and query pairs $\mathbb{T}$ while our additional $R_{pir}$ ensures that the MLLM can also get rewards in condition that the MLLM refuses to respond if it is unable to find any semantic match for the irrelevant video query pairs $\mathbb{F}$. Noticeably, the proportion of irrelevant to relevant pairs is crucial according to our following ablation experiment.
\begin{equation}
\label{eq5}
\begin{split}
R_{p} &= R_{tvg} + R_{pir} \\
&=\underset{(i\in\mathbb{T})} {||\sum\frac{Prd_{i}\cap GT_{i}}{Prd_{i}\cup GT_{i}}||_{norm}} +\underset{(i\in\mathbb{F})}{||\gamma\cdot(\sum\mathbb{I}_{Prd_{i}=\emptyset})||_{norm}}
\end{split}
\end{equation}
Therefore, the final reward function of \textbf{PIR-GRPO} is defined as the weighted combination of $R_{p}$ and $R_{f}$ in \eqref{eq6}:
\begin{equation}
\label{eq6}
R = \alpha\cdot R_{f} + \beta\cdot R_{p}
\end{equation}

\subsection{Testbench with More Reasonable Ground Truth (GT) Annotations}
\begin{figure*}[!ht]
\centering
\captionsetup{justification=raggedright, singlelinecheck=false}
\includegraphics[width=1.97\columnwidth]{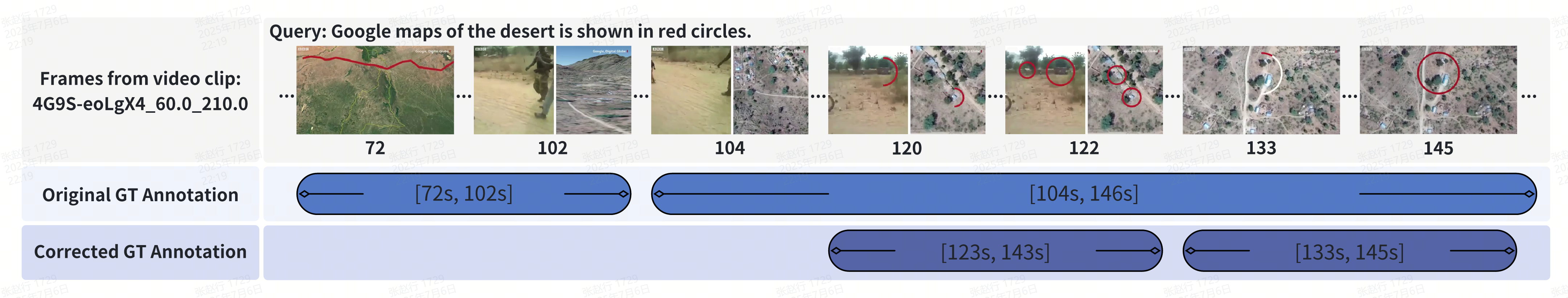}
\caption{One of the inappropriate GT annotation example in the QvHighlights testbench with our corrected annotation.}
\label{tr0-i3}
\end{figure*}

We observe inaccuracies in the GT boundary annotations of query-relevant video clips in QvHighlight dataset. To mitigate this issue, we manually rectify these erroneous annotations and provide a corrected testbench dubbed as cQvH as Figure \ref{tr0-i3} display. For the fair comparison with other SOTA methods, either QvH or CQvH testbench will be utilized in the following experiments.

\section{Experiments}
\begin{table*}[!ht]
\centering
\caption{Comparison results on the several datasets.} 
\scalebox{1.0}{
\renewcommand{\arraystretch}{1.29}
\begin{tabular}{p{2.6cm}>{\centering\arraybackslash}p{2.6cm}>{\centering\arraybackslash}p{.8cm}>{\centering\arraybackslash}p{.8cm}>{\centering\arraybackslash}p{.8cm}>{\centering\arraybackslash}p{.8cm}>{\centering\arraybackslash}p{.8cm}>{\centering\arraybackslash}p{.8cm}>{\centering\arraybackslash}p{.8cm}>{\centering\arraybackslash}p{.8cm}
}
\midrule
\midrule
\multirow{4}{*}{\textbf{Solutions}} & \multirow{4}{*}{\textbf{(M)LLMs}} & \multicolumn{8}{c}{\textbf{Datasets}} \\ 
\cmidrule{3-10}
 &  & \multicolumn{3}{c}{\textbf{QvH}} & \multicolumn{2}{c}{\textbf{C-STA}}  & \multicolumn{3}{c}{\textbf{cQvH}} \\ 
\cmidrule(r){3-5} \cmidrule(r){6-7} \cmidrule(r){8-10}
 & & \textbf{R@.5} & \textbf{R@.7} & \textbf{mAP} & \textbf{R@.5} & \textbf{R@.7} & \textbf{R@.5} & \textbf{R@.7} & \textbf{mAP}\\
\midrule
Chrono & BLIP-2 & 74.77 & 60.51 & 51.37 & 69.31 & 49.29 & 79.68 & 67.87 & 57.97\\
SeViLa & BLIP-2 & 54.50 & 36.50 & 32.30 & - & - & - & - & -\\
InternVideo2-6B & BLIP-2 & 71.42 & 56.45 & 49.24 & 70.03 & 48.95 & - & - & -\\
Tempo-R0 & Qwen2-VL-7B & \textcolor{red}{78.52} & \textcolor{red}{65.23} & 54.50 & 69.73 & 45.13 & \textcolor{red}{84.65} & \textcolor{red}{72.39} & \textcolor{red}{61.47}\\
\midrule
\midrule
\end{tabular}
}
\label{tr0-t0}
\end{table*}

\subsection{\textbf{Implementation Details}}
The backbone MLLM of Tempo-R0 is Qwen2-VL-7B, with both text and visual encoders using its built-in configurations. First, we initialize Qwen2-VL-7B with pretrained parameters, followed by fine-tuning it via \textbf{PIR-GRPO}. In the first stage, we conduct SFT using the training set parts of QvHightlight, Charades-STA\cite{Gao_2017} and add some data from the OOD dataset InternVid\cite{wang2022internvideo} to enhance the basic temporal reasoning capability of Qwen2-VL-7B. However, similarity tests revealed that this OOD dataset is only similar in content to QvHighlights. Thus, we perform data augmentation solely for QvHighlights. In the second stage, we only use the ID training sets mixed with about ten-percent of augmented irrelevant video query training pairs.

\subsection{\textbf{Data and Evaluation}}
We validate Tempo-R0 on the following mainstream TVG task datasets:
\begin{itemize}
    \item \textbf{QvHighlights:} This dataset primarily comprises lifestyle vlogs sourced from user-generated content on YouTube.
    \item \textbf{Charades-STA:} This dataset focus on various human activities within indoor environments.
    \item \textbf{ActivityNet \cite{anet2015Heilbron}:} A Dataset for dense video captioning containing multiple sentence-moment pairs applicable to TVG Tasks.
\end{itemize}

\subsection{\textbf{Main Results}}
As illustrated in Table \ref{tr0-t0}, we compare our method to other SOTA Video-MLLM approaches \cite{meinardus2024chrono, Wang_2024, yu2023selfchained} in QvHighlights (QvH), corrected QvHighlights (cQvH), Charades-STA (C-STA) and ActivityNet (ANet). On the QvH and cQvH testbenchs, our method shows significant advantages over others. For C-STA, Tempo-R0 also achieves decent performance. These results demonstrate that Tempo-R0 exhibits robust temporal reasoning capabilities by means of our amelioration. We only compare with those methods that provide the statistics on the QvH testbench. 

We also employ transfer learning to investigate the generalization extent of the model's TVG capability. Specifically, we fine-tune the model on the QvH training set and evaluate its performance in a zero-shot manner on the ANet test dataset and we also only list the comparison with the solutions that perform this experiment \cite{li2025videochatr1, wang2025timer1}.

\begin{table}[!ht]
\centering
\caption{Transfer Learning Comparison.} 
\scalebox{1.0} {
    \renewcommand{\arraystretch}{1.29}
    \begin{tabular} {lcc}
    \midrule
    \midrule
    \multirow{2.5}{*}{\textbf{Solutions}} & \multicolumn{2}{c}{\textbf{QvH} $\longrightarrow$ \textbf{ANet}}
    \\ 
    \cmidrule(r){2-3} & \textbf{R@.5} & \textbf{R@.7}
    \\
    \midrule
    BLIP-Chrono & 28.64 & 16.44 \\
    VideoChat-R1 & 32.2 & 16.2 \\
    Tempo-R1 & \textcolor{red}{32.22} & \textcolor{red}{18.27} \\
    \midrule
    \midrule
    \end{tabular}
}
\label{tr0-t2}
\end{table}
The comparison result in Table \ref{tr0-t2} also shows our Tempo-R0's advantage.

\subsection{Ablation Experiments}
\begin{figure*}[!ht]
\centering
\captionsetup{justification=raggedright, singlelinecheck=false}
\includegraphics[width=1.97\columnwidth]{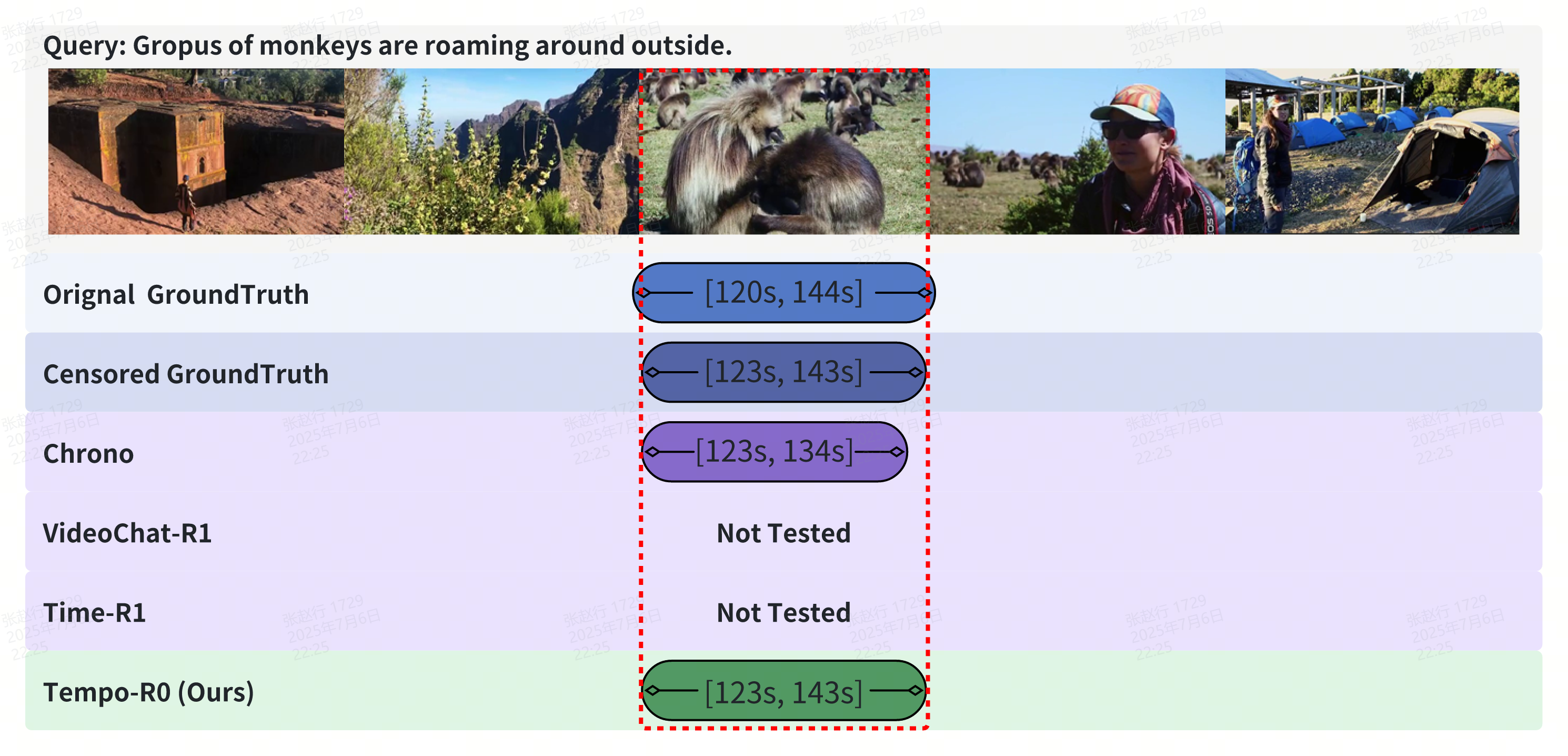}
\caption{The comparison of a VTG example in QvH and cQvH testbenchs. }
\label{tr0-i4}
\end{figure*}

A detailed ablation on the Tempo-R0 model to investigate the contribution of our designed modules and exploration. And we select QvH test dataset to perform these ablation experiments.

We first list the comparison results in Table \ref{tr0-t3} and the result demonstrates an accuracy gain with our \textbf{SAA} module.
\begin{table}[!ht]
\centering
\caption{Ablation experiment on \textbf{SAA}} 
\scalebox{1.0} {
    \renewcommand{\arraystretch}{1.29}
    \begin{tabular} {lccc}
    \midrule
    \midrule
    & \textbf{R@.5} & \textbf{R@.7} & \textbf{mAP}\\ 
    \midrule
    w/ \textbf{SAA} & \textcolor{red}{77.39} & \textcolor{red}{63.52} & \textcolor{red}{53.84} \\
    w/o \textbf{SAA} & 76.77 & 63.61 & 53.33 \\
    \midrule
    \midrule
    \end{tabular}
}
\label{tr0-t3}
\end{table}

The application of \textbf{SAA} introduces more visual tokens. We thus conduct another ablation study on the impact of visual token number and the result in Table \ref{tr0-t5} indicates that the increase in tokens does not necessarily lead to improved TVG accuracy, which makes \textbf{SAA} the contributing factors of the gain.
\begin{table}[!ht]
\centering
\caption{Ablation experiment on token number} 
\scalebox{1.0} {
    \renewcommand{\arraystretch}{1.29}
    \begin{tabular} {cccc}
    \midrule
    \midrule
    \textbf{Visual Tokens} & \textbf{R@.5} & \textbf{R@.7} & \textbf{mAP}\\ 
    \midrule
    6000 & \textcolor{red}{72.71} & \textcolor{red}{55.03} & \textcolor{red}{45.38} \\
    12000 & 59.81 & 43.48 & 43.13 \\
    20000 & 67.16 & 48.71 & 38.33 \\
    30000 & 68.45 & 51.16 & 42.74 \\
    \midrule
    \midrule
    \end{tabular}
}
\label{tr0-t5}
\end{table}

In the following ablation experiment for \textbf{ETA}, its effectiveness can also be observed from \ref{tr0-t4}.
\begin{table}[!ht]
\centering
\caption{Ablation experiment on \textbf{ETA}} 
\scalebox{1.0} {
    \renewcommand{\arraystretch}{1.29}
    \begin{tabular} {lccc}
    \midrule
    \midrule
    & \textbf{R@.5} & \textbf{R@.7} & \textbf{mAP}\\ 
    \midrule
    w/ \textbf{ETA} & \textcolor{red}{76.77} & \textcolor{red}{63.61} & \textcolor{red}{53.33} \\
    w/o \textbf{ETA} & 76.13 & 61.94 & 50.22 \\
    \midrule
    \midrule
    \end{tabular}
}
\label{tr0-t4}
\end{table}

The \textbf{PIR-GRPO} ablation experiment result in Table \ref{tr0-t6} demonstrates that the adoption of \textbf{PIR-GRPO} leads to a significant improvement in TVG accuracy than other manipulations such as the addition of OOD in SFT stage. Furthermore, we list several results under different ratios of irrelevant video-query pairs in the entire training set and the comparison in Table \ref{tr0-t7} shows that around 10\% ratio yields the averagely best results.
\begin{table}[!ht]
\centering
\caption{Ablation experiment on \textbf{PIR-GRPO}} 
\scalebox{1.0} {
    \renewcommand{\arraystretch}{1.29}
    \begin{tabular} {lccc}
    \midrule
    \midrule
    Fine-tuning Options & \textbf{R@.5} & \textbf{R@.7} & \textbf{mAP}\\ 
    \midrule
    Baseline with SFT & 73.35 & 58.97 & 50.53 \\
    +OOD & 74.06 & 60.0 & 52.1 \\
    +OOD +\textbf{PIR-GRPO} & \textcolor{red}{78.52} & \textcolor{red}{65.23} & \textcolor{red}{54.50} \\
    \midrule
    \midrule
    \end{tabular}
}
\label{tr0-t6}
\end{table}
\begin{table}[!ht]
\centering
\caption{Irrelevance Ratio Comparison.} 
\scalebox{0.9} {
    \renewcommand{\arraystretch}{1.29}
    \begin{tabular} {ccccc}
    \midrule
    \midrule
    \multirow{2}{*}{\textbf{Irrelevance Paris Ratio}} & \multicolumn{3}{c}{\textbf{$\mathbb{T}$ Dataset}} & \multicolumn{1}{c}{\textbf{$\mathbb{F}$ Dataset}} \\ 
    \cmidrule(r){2-4} \cmidrule(r){5-5} & \textbf{R@.5} & \textbf{R@.7} & \textbf{mAP} & \textbf{R} \\ 
    \midrule
    QvH(0\%) & 73.35 & 58.97 & 50.53 & 0.1 \\
    QvH(5\%) & 76.19 & 62.45 & 52.76 & 86.6 \\
    QvH(10\%) & \textcolor{red}{78.84} & \textcolor{red}{64.65} & \textcolor{red}{54.22} & 81.7 \\
    QvH(20\%) & 76.26 & 63.68 & 53.14 & \textcolor{red}{94.6} \\
    QvH(30\%) & 77.16 & 63.68 & 53.1 & 85.5 \\
    \midrule
    \midrule
    \end{tabular}
}
\label{tr0-t7}
\end{table}

\subsection{Case Study}
We show an example that Tempo-R0 outperformes other methods in QvH and cQvH testbench in Figure \ref{tr0-i4}.

\section{Conclusion}
In this work, we propose an efficient video-MLLM-based solution Tempo-R0 for TVG task featuring strong temporal reasoning capability with the aid of our multimodal temporal reasoning strengthening. We present that Tempo-R0 achieves SOTA performance in the mainstream testbench. Through ablation studies, we further prove that these creative components like \textbf{SAA}, \textbf{ETA} and \textbf{PIR-GRPO} of Tempo-R0 all contribute to the model's overall temporal reasoning capability.

\newpage
\newpage
\bibliography{aaai2026}

\end{document}